\documentclass{llncs}

\usepackage{amsmath,amssymb,amsfonts}
\usepackage{graphicx}
\usepackage[draft]{fixme}
\usepackage[table]{xcolor}

\usepackage[pagebackref]{hyperref} % pagebackref option links reference entries back to the page of the citation
\hypersetup{colorlinks=true} % if false, use colored frames that do not print
\usepackage[boxruled]{algorithm2e} % norelsize is needed when using amsart document class to avoid error
\usepackage{paralist}                           % for inparaemun
\usepackage{bm}
\usepackage{float}
\usepackage{multirow}
\usepackage{booktabs}
\usepackage{nicefrac}
\usepackage{nth}
\usepackage[bf,small,skip=0pt]{caption}
\usepackage{subcaption}
\usepackage{wrapfig}
\usepackage{dsfont}
\usepackage{tikz}
\usetikzlibrary{shapes,arrows,backgrounds,positioning}
\usepackage{amssymb}
\usepackage{amsmath}

\usepackage{mathtools}
\usepackage{dsfont}

\DeclareMathOperator*{\argmax}{arg\,max}

\usepackage{caption}
\usepackage{adjustbox}
\usepackage{multirow}

\usepackage{url}
\urldef{\mailsa}\path|{zp-ding, mn}@cs.unc.com|

\usepackage{color}

\usepackage{wrapfig}

\begin{document}

\mainmatter  % start of an individual contribution

\title{VoteNet: A Deep Learning Label Fusion Method for Multi-Atlas Segmentation}

\titlerunning{VoteNet: A Deep Learning Based Label Fusion Method for Multi-Atlas Segmentation}

\author{Zhipeng Ding$^1$, Xu Han$^1$, Marc Niethammer$^{1,2}$}
%index{Ding, Zhipeng}
% index{Fleishman, Greg}
%index{Xu Han}
% index{Thompson, Paul}
% index{Kwitt, Roland}
%index{Niethammer, Marc}
%\author{Zhipeng Ding%
%%\thanks{Please note that the LNCS Editorial assumes that all authors have used
%%the western naming convention, with given names preceding surnames. This determines
%%the structure of the names in the running heads and the author index.}%
%\and Greg Fleishman\and Xiao Yang\and Paul Thompson\and Roland Kwitt\and Marc Niethammer
%\and The Alzheimer's Disease Neuroimaging Initiative}
%%
%\authorrunning{MICCAI paper preparation}

%\institute{Department of Computer Science,\\
%University of North Carolina at Chapel Hill, USA\\
\institute{$^1$Department of Computer Science, UNC Chapel Hill, USA\\
$^2$Biomedical Research Imaging Center, UNC Chapel Hill, USA\\}
%$^3$Imaging Genetics Center, USC, USA\\
%$^4$Department of Radiology, University of Pennsylvania, USA\\
%$^5$Department of Computer Science, University of Salzburg, Austria}
%University of North Carolina at Chapel Hill, USA\\
%\url{http://www.springer.com/lncs}
%}


\maketitle

\begin{abstract}

  Deep learning (DL) approaches are state-of-the-art for many medical image segmentation tasks. They offer a number of advantages: they can be trained for specific tasks, computations are fast at test time, and segmentation quality is typically high. In contrast, previously popular multi-atlas segmentation (MAS) methods are relatively slow (as they rely on costly registrations) and even though sophisticated label fusion strategies have been proposed, DL approaches generally outperform MAS. In this work, we propose a DL-based label fusion strategy (VoteNet) which locally selects a set of reliable atlases whose labels are then fused via plurality voting. Experiments on 3D brain MRI data show that by selecting a good initial atlas set MAS with VoteNet significantly outperforms a number of other label fusion strategies as well as a direct DL segmentation approach. We also provide an experimental analysis of the upper performance bound achievable by our method. While unlikely achievable in practice, this bound suggests room for further performance improvements. Lastly, to address the runtime disadvantage of standard MAS, all our results make use of a fast DL registration approach.

\end{abstract}

\section{Introduction}
Image segmentation, i.e. giving pixels or voxels in an image meaningful labels, is an important medical image analysis task~\cite{iglesias2015multi}. While a large number of segmentation approaches exist, deep convolutional neural networks (CNNs) have shown remarkable segmentation performance and are considered state-of-the-art~\cite{cciccek20163d,long2015fully}. This was not always the case. Prior to the dominance of CNNs, multi-atlas segmentation (MAS) has been highly popular and successful for medical image segmentation~\cite{iglesias2015multi}. MAS approaches rely on a set of previously labeled atlas images. These images are then registered to an unlabeled target image and their associated labels are used to infer the labeling of the target image via label fusion. Hence, MAS performance relies on high-quality registrations or advanced label fusion methods. %Label fusion strategies help account for misregistrations or unsuitable atlases; they resolve local label inconsistencies between the warped atlas labels.
MAS is slow as it requires computationally costly registrations, but provides good spatial consistency via the given atlas segmentations.\\

  In contrast, CNN approaches use sophisticated network architectures, with parameters trained on large sets of labeled images. A popular architecture for medical image segmentation is the U-Net~\cite{cciccek20163d}. For DL approaches, the majority of computational cost is spent during training. Hence, these approaches are fast at test time and have shown excellent segmentation performance for medical images~\cite{cciccek20163d,long2015fully}. However, as image labels are not directly spatially transformed, spatial consistency is only indirectly encouraged during training and DL approaches may miss or add undesired structures. Furthermore, large numbers of labeled images are desirable for training, but may not always be available.\\

  Conceptually, MAS is attractive as it provides a direct and intuitive way to specify and obtain segmentations via a set of labeled atlases. Given atlases that are similar to the target image to be segmented, it is plausible that good registrations can be achieved, that atlas labels can be transferred well, and consequentially that high-quality segmentations can be obtained. However, it is a-priori unclear which atlases should be used to estimate the segmentation as not all atlases will align well via registration. Label fusion strategies aim at addressing the resulting spatial inconsistencies between the atlas labels warped to the target image space. Approaches include, majority and plurality voting~\cite{heckemann2006automatic,hansen1990}, global weighted voting~\cite{artaechevarria2008efficient}, and local weighted voting strategies~\cite{wang2013multi}. Statistical modeling approaches have also been proposed~\cite{warfield2004simultaneous} and patch-based approaches directly aim to compensate for local registration errors~\cite{bai2013probabilistic}. Most recently, machine learning~\cite{wang2014multi} and deep learning~\cite{yang2018neural} label fusion methods have been proposed. These methods all assume that all atlases might contribute to the labeling decision. Instead, we propose making decisions only based on atlases considered trustworthy. In contrast to global~\cite{sanroma2014learning} and patch-based atlas selection~\cite{konukoglu2013neighbourhood}, our approach {\it locally} predicts the set of trustworthy atlases voxel by voxel. %\mnl{This needs to be checked. I was under the impression, from the way this was written, that there are no local atlas selection strategies. If there are, they need to be discussed, since this is exactly what they are doing. Obviously one would need to compare to these methods (if they exist), but it is too late for that now.\zpd{(The only local atlas selection that I found is the patch-based method. It is not exactly a MAS setting because the author did not do any registration to align the images. They just divide the target image into overlapping patches and for each patch select the closest 20 patches out of 30,000 training patches using their method called NAF. And then use NCC to refine the 20 patches in a search window. Finally they majority vote on overlaps. The patch size is $36\times36\times36$. I would argue that this is still a semi-global atlas selection. Ours is truly local atlas selection and we only consider the fixed number fixed position voxels. Our motivation is the Oracle results which convey the information that knowing which atlas is trustworthy is very improtant.)}}
Experimentally, we show that this strategy, even combined with simple plurality voting, has excellent segmentation performance on par or slightly outperforming a U-Net and significantly outperforming other label fusion strategies. All our results make use of fast DL-based deformable image registration, thereby resulting in a MAS approach which is fast and accurate while providing more direct control over spatial label consistency.\\

  \textbf{Contributions.} (1)~\emph{New label fusion method}: We propose a DL label fusion method (VoteNet), which locally identifies sets of trustworthy atlases. (2)~\emph{Fast implementation}: All our results are based on a fast DL registration approach (Quicksilver). (3)~\emph{Performance upper bound}: We experimentally assess the best possible performance achievable with our approach and illustrate that there is a large margin for improvements. (4)~\emph{Comprehensive experimental comparison}: We compare to a variety of other label fusion strategies and a U-Net for 3D brain segmentation. Our approach performs consistently best.\\

%(2)~\emph{Fast image registration prediction}: We employed a fast predictive image registration method in our MAS framework to overcome the computational bottleneck. (3)~\emph{Fully deep learning based MAS pipeline}: We proposed to use one registration network and one fusion network to reformulate MAS problem; this skeleton simplifies MAS into the well-studied segmentation problem.  

\textbf{Organization.} Sec.~\ref{sec:method} describes our proposed MAS framework in detail. Sec.~\ref{sec:experiments_and_results} discusses experimental details and results on the LPBA40 dataset. Sec.~\ref{sec:conclusion} concludes the paper with a summary and an outlook on future work. %\zpd{(maybe delete the organization part to save some space.)}

\section{Methodology}
\label{sec:method}

\begin{figure}[t]
  \includegraphics[width=\linewidth]{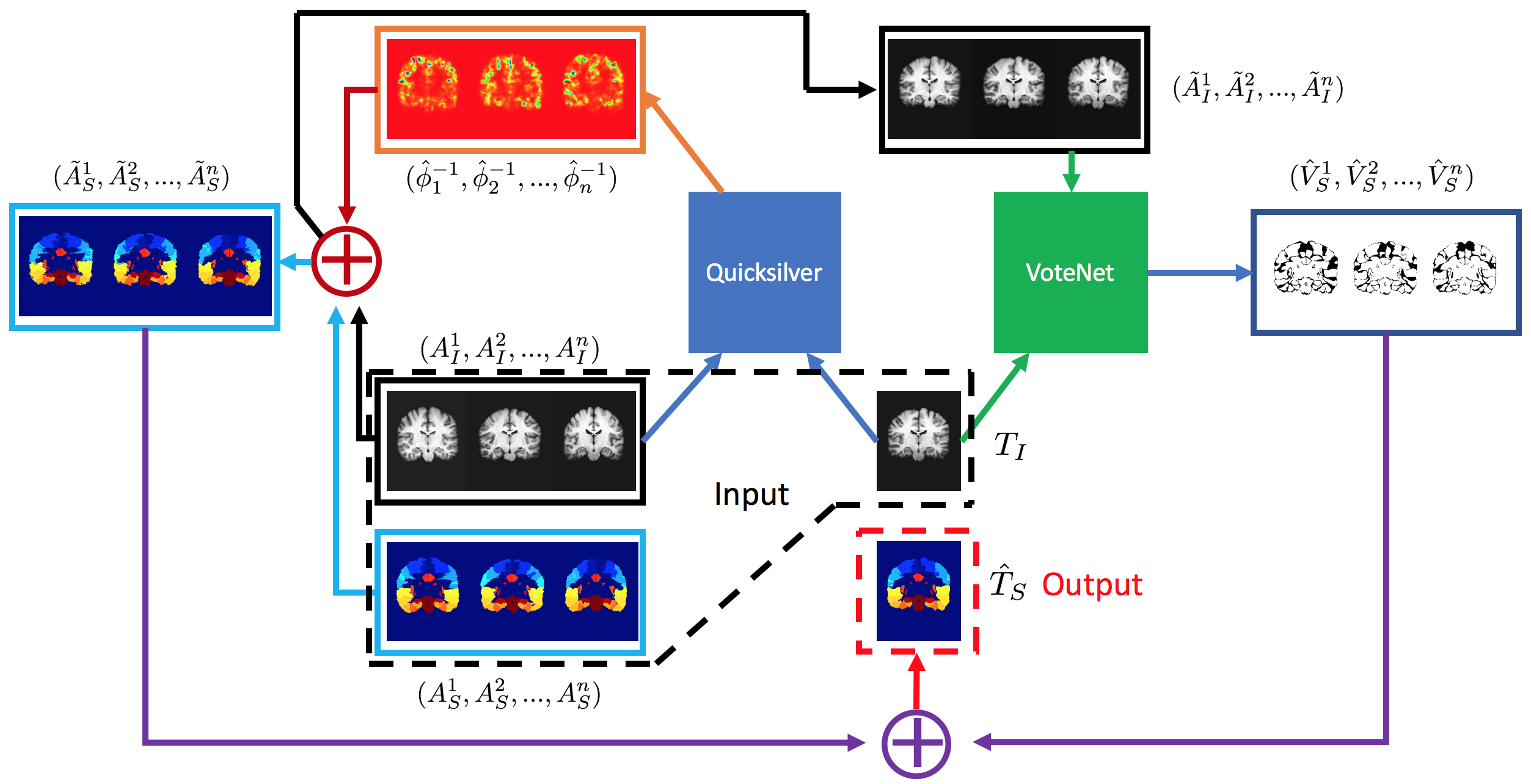}
  \caption{Our MAS framework. Atlas images and target image are input into Quicksilver to predict transformation maps $\hat{\phi}^{-1}$ from atlas to target space. The atlas images and labels are then warped to target space by $\hat{\phi}^{-1}$. VoteNet uses the warped atlas images and the target image to predict a binary mask $\hat{V}_S$ for each atlas indicating locally if an atlas should be used. Final label fusion is done via plurality voting using only the atlases which should be used according to the VoteNet prediction. See Eq.~\ref{eq:goal}.}
  \label{fig:framework}
\end{figure}

Before discussing our proposed label fusion method, we first describe multi-atlas segmentation. Let $T_I$ be the target image to be segmented and $A^1 = (A^1_I, A^1_S), A^2 = (A^2_I, A^2_S), ..., A^n = (A^n_I, A^n_S)$ be $n$ atlas images and their corresponding manual segmentations. Assume there is a reliable deformable image registration method that warps all atlases into the space of the target image $T_I$, i.e. $\tilde{A}^1 = (\tilde{A}^1_I, \tilde{A}^1_S), \tilde{A}^2 = (\tilde{A}^2_I, \tilde{A}^2_S), ..., \tilde{A}^n = (\tilde{A}^n_I, \tilde{A}^n_S)$. Each $\tilde{A^i_S}$ is now a candidate segmentation for $T_I$. Finally, a label fusion method $\mathcal{F}$ combines all the candidate segmentations to produce the final segmentation $\hat{T}_S$ for $T_I$, i.e.,
\begin{equation}
\label{eq:est}
\hat{T}_S = \mathcal{F}(\tilde{A}^1, \tilde{A}^2, ..., \tilde{A}^n, T_I).
\end{equation}

Our framework uses two deep convolutional neural networks (CNNs) for MAS: the Quicksilver registration network to compute the spatial transformation to target image space and our label fusion network (VoteNet). While other registration approaches could be used, a DL approach greatly speeds-up the typically slow registrations for MAS, when based on numerical optimization. By using a DL approach we also demonstrate that it integrates well with MAS.\\ %The label fusion network is used to predict the final labels based on warped labels.

Fig.~\ref{fig:framework} illustrates our approach. Quicksilver and VoteNet are discussed below.

\noindent \textbf{Quicksilver:} Quicksilver~\cite{quicksilver} uses the target image $T_I$ and the atlas images $(A^1_I, A^2_I, ..., A^n_I)$ to predict the deformation maps $(\hat{\phi}^{-1}_1, \hat{\phi}^{-1}_2, ..., \hat{\phi}^{-1}_n)$ which are used to generate warped atlas images $(\tilde{A}^1_I, \tilde{A}^2_I, ..., \tilde{A}^n_I)$ and their corresponding labels $(\tilde{A}^1_S, \tilde{A}^2_S, ..., \tilde{A}^n_S)$. Using a DL registration approach such as Quicksilver speeds-up pairwise registrations by at least an order of magnitude~\cite{quicksilver} compared to numerical optimization. Since registrations are the computational bottleneck of MAS, similar speed-ups can be obtained. E.g., MAS with our approach and 17 atlases requires only 15 mins on an NVIDIA GTX1080Ti. Experiments (Sec.~\ref{sec:experiments_and_results}) show that Quicksilver yields good results when combined with MAS (Tab.~\ref{tab:metrics}). \\%In Tab.~\ref{tab:metrics}, We achieved an averaged volume dice score 80.55, which is slightly better than U-Net.} \mnl{Zhipeng, give some example results here.\zpd{(done)}}

%The detail of Quicksilver can be found in the paper~\cite{quicksilver}. We used all 373 images from the OASIS longitudinal dataset as the training data, and randomly select target images from different subjects for every image, creating 373 registrations for the training of the network. We replicated exactly the author's strategy to train the quicksilver registration network. Details can be found in the github\footnote{https://github.com/rkwitt/quicksilver}.\\

\noindent
\textbf{VoteNet:} Given the warped atlas images $(\tilde{A}^1_I, \tilde{A}^2_I, ..., \tilde{A}^n_I)$ and the target image, VoteNet independently predicts binary masks $(\hat{V}^1_S, \hat{V}^2_S, ..., \hat{V}^n_S)$ for each warped atlas image, locally indicating if a warped atlas should be considered for the final labeling decision of the target image. In effect, VoteNet predicts for each spatial location the \emph{set of trustworthy atlases};  all atlases which should likely not be used are discarded. Hence, VoteNet, implements a form of locally adaptive statistical trimming. Once the set of trustworthy atlases has been determined, their associated labels can be fused with any chosen label fusion strategy. For simplicity we use plurality voting. Our VoteNet strategy shifts the notion of a plurality to a \emph{plurality of trusted atlases}. We define \emph{trusted plurality voting} as
\begin{equation}
\label{eq:goal}
\hat{T}_S(x) = \argmax_{l \in \Omega} \sum_{i=1}^{n} \mathds{1}[\hat{V}^i_S(x) \odot \tilde{A}^i_S(x) = l],
\end{equation}
where $l \in \Omega=\{0,\dots,N\}$ is the set of labels ($N$ structures; 0 indicating background), $\mathds{1}[\cdot]$ is the indicator function and $x$ denotes a voxel position. We define $\hat{V}^i_S (x) \odot \tilde{A}^i_S (x) := \tilde{A}^i_S (x)\in\Omega$ if $\hat{V}^i_S (x) = 1$; $\hat{V}^i_S (x) \odot \tilde{A}^i_S (x) := -1$ if $\hat{V}^i_S (x) = 0$.\\

%Note that we just used simple majority voting after masking as the label fusion method. The performance is already really well. More advanced label fusion methods are left as future research.\\

%~\cite{marcus2007open}

%One major problem of MAS is that some voxels in warped atlases are not correct. Thus, we designed a CNN to do a binary prediction voxel-wisely.

\begin{figure}[t]
  \includegraphics[width=\linewidth]{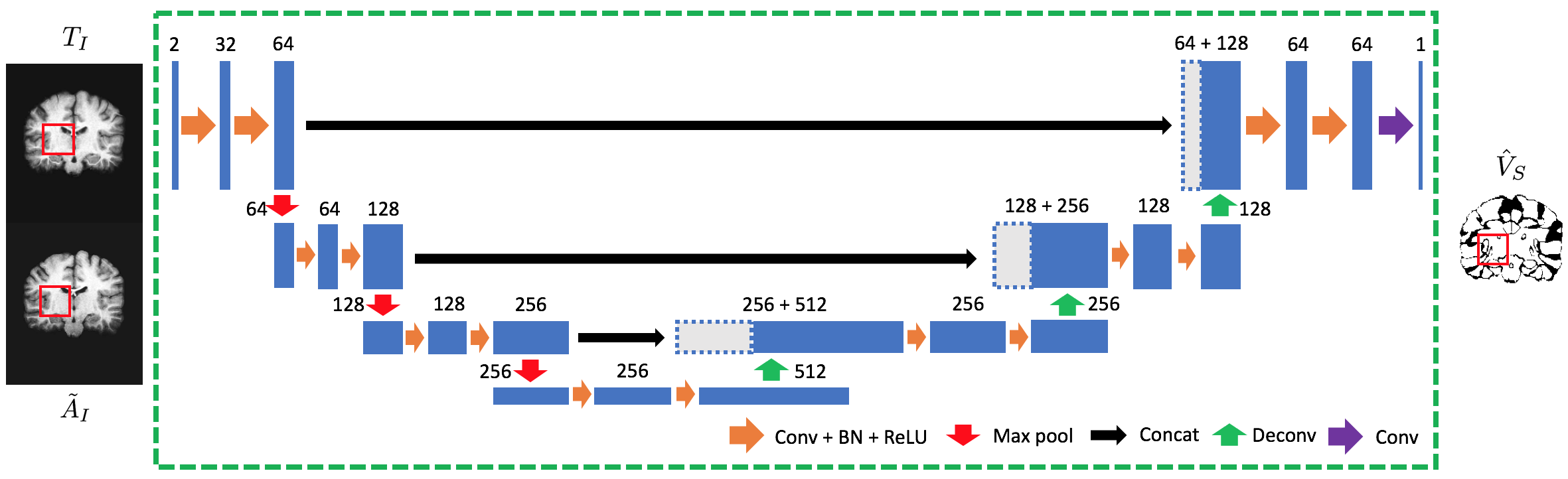}
  \caption{VoteNet architecture. We use a U-Net structure with skip connections and batch normalization. Numbers indicate feature numbers (channels) in each convolutional layer. Inputs are target image and a warped atlas image. Output is a binary mask (after thresholding the sigmoid output at 0.5) to select trustworthy atlas voxels.}
  \label{fig:votenet}
\end{figure}

\noindent
\textbf{VoteNet training:} Fig.~\ref{fig:votenet} shows the VoteNet architecture, which is based on the 3D U-Net~\cite{cciccek20163d} (but uses the target image and a warped atlas image as inputs). VoteNet processes an image patch-wisely, with a patch size of $72\times72\times72$ from the target image $T_I$ and a warped atlas image $\tilde{A}_I$ at the same position, where the $40\times40\times40$ patch center is used to tile the volume. %\mnl{How do you combine the patches\zpd{(If you mean how do I combine the patches into the network, I just concatenate the two patches in the channel; if you mean in testing time, if I have a overlap of 16, so the effective size for a patch is actually $40\times40\times40$. Then the image size divide the effective size to determine the grid size. The final padding size is calculated by grid size $\times$ effective size + overlap size $\times$ 2 - image size. I use overlapping tiles as Unet did.)}}
To train VoteNet, we use 20 images from LONI Probabilistic Brain Atlas dataset (LPBA40)\footnote{LPBA40 contains 40 3D brain MRIs with 56 manually segmented structures. Preprocessing includes affine registration to the MNI152 atlas and histogram equalization.}; we randomly select 17 images and their labels as atlases and use Quicksilver to register all 17 atlases to the 20 images excluding themselves. This results in 323 ($17 \times 16 + 17 \times 3$) pairwise registrations. Given the manual segmentation labels of these 20 images, we determine at which location the warped atlas labels agree (1) and where they disagree (0). These agreement/disagreement labels are our training labels for VoteNet, trained via a binary cross entropy loss (Fig.~\ref{fig:votenet}) in PyTorch. As VoteNet produces continuous outputs, $P_i$, (in $(0,1)$ via a sigmoid) we threshold at 0.5 at test time, i.e., the local prediction is $\hat{V}^i_S = 0$ if $P_i < 0.5$ and $\hat{V}^i_S = 1$ otherwise. We train using \texttt{ADAM} over 300 epochs with a multi-step learning rate. The initial learning rate is 0.001; reduced to 0.0001 after 150 epochs; and finally to 0.00001 after 250 epochs.%\mnl{Too late now, but what motivates this somewhat strange strategy? Why not simply use a learning rate scheduler as available in PyTorch.\zpd{(I am using the learning rate scheduler in PyTorch (lr\_scheduler.MultiStepLR). I just customized the number of epochs that I want to use the learning rate. This is based on observation of the loss function and validation performance to set when to use a lower learning rate.)}}
Training image patches were randomly extracted so that 0 labels account for no less than 5\% of the entire patch volume. %\mnl{How exactly is this done.\zpd{(Randomly crop the patch, then calculate the percentage of 0 label out of the patch area. If it is more than 5\%, then use it otherwise randomly crop another patch until meet this criteria.)}}
Training requires $\approx$24h. The prediction of a single atlas mask takes $<$20s. 

%The output is a binary mask (after sigmoid and thresholding at 0.5) of the same size as input patch size that indicates whether the corresponding voxel is correct or not.

%\section{Implementation Details}
%\label{sec:implementation_detail}
%We train our VoteNet using \texttt{ADAM} over 300 epochs with multi-step learning rate. The initial learning rate was 0.001; after 150 epochs it became 0.0001; finally it became 0.00001 after 250th epoch. We extracted patches at a ratio that the non-voting voxels account no less than 5\% of the entire patch volume. The training process requires around 24 hours on a NVIDIA GTX 1080Ti, while the prediction of a single atlas mask takes less than 20 seconds.

%Binary cross entropy (BCE) loss is used as our loss function:
%\begin{equation}
%\label{eq:loss}
%\mathcal{L}_{BCE} = - V^i_S \cdot \ln(P_i) + (1-V^i_S) \cdot \ln(1 - P_i)
%\end{equation}
%where the output of the VoteNet, $P_i$, is the probability of class $i$ ($i$ = 0 or 1), $V_S^i$ is the ground truth binary mask in which 0 means the voxel is incorrect and 1 means the voxel is correct. When testing, the prediction $\hat{V}^i_S = 0$ if $P_i < 0.5$, otherwise $\hat{V}^i_S = 1$.

\section{Experimental Results and Discussion}
\label{sec:experiments_and_results}
We use LPBA40 for evaluation. We use two-fold cross-validation, i.e. the dataset is randomly divided into two non-overlapping subsets of equal size. One set is used for training (Sec.~\ref{sec:method}) the other for testing for each of the two cross-validation experiments. The results below are averaged over the cross-validation folds.\\

%In each training set, we randomly select 17 out of 20 images as atlases for predicting the segmentations in the testing set and the rest 3 images as the validation set for fine-tuning the model parameters. 

%~\cite{shattuck2008construction}, ~\cite{grabner2006symmetric}

\noindent
\textbf{Benchmark methods:} We compare against plurality voting (PV)~\cite{heckemann2006automatic}, majority voting (MV), simultaneous truth and performance level estimation (STAPLE)~\cite{warfield2004simultaneous}, multi-atlas based multi-image segmentation (MAMBIS)\footnote{MABMIS uses Diffeomorphic Demons for registration and hence results are not directly comparable to ours; we use Quicksilver for all other label fusion methods.}~\cite{jia2012iterative}, joint label fusion (JLF)~\cite{wang2013multi}, patch-based label fusion (PB)~\cite{bai2013probabilistic} and a U-Net~\cite{cciccek20163d}. PV locally assigns the most frequent segmentation label among the atlases. MV assigns a label only if more than half of the atlases ($\ge 9$) agree. STAPLE uses a statistical model to estimate a true hidden segmentation based on an optimal weighting of the segmentations. MAMBIS puts atlases in a tree structure to consider their correlations for concurrent alignment. JLF regards label fusion as an optimization problem and minimizes the total expectation of labeling errors. PB searches in a neighborhood to reduce registration errors and utilizes patch intensity and label information within a Bayesian label fusion framework. %\zpd{The patch size is $7 \times7\times7$ and search volume size is $5\times5\times5$}.
We include a U-Net for a direct comparison to a popular DL image segmentation approach. We also create an oracle label fusion strategy, which has access to the true label during local atlas selection. This allows establishing upper performance bounds.\\

\begin{table*}[!t] 
\centering
\normalsize
\begin{adjustbox}{max width=\textwidth}
%\begin{tabular}{ |c|p{2cm}|p{2cm}|p{2cm}|p{2cm}|p{2cm}|p{2cm}|p{2cm}|}
\begin{tabular}{ |c|c|c|c|c|c|}
\hline  
Method & Avg. Surf. Dist. (mm)& Surf. Dice (\%) & Hausdorff Dist. (mm)& 95\% Max. Dist. (mm)& Volume Dice (\%)\\ 
\hline 
Oracle(1)&0.04 $\pm$ 0.01&99.06 $\pm$ 0.23 & 3.64 $\pm$ 0.32& 0.18 $\pm$ 0.08&98.99 $\pm$ 0.19\\ 
\hline
Oracle(3)&0.17 $\pm$ 0.03& 95.91 $\pm$ 0.66& 5.80 $\pm$ 0.43& 1.17 $\pm$ 0.16&96.45 $\pm$ 0.46 \\
\hline
Oracle(6)&0.45 $\pm$ 0.05&88.97 $\pm$ 1.17 &8.00 $\pm$ 0.51& 2.46 $\pm$ 0.22&91.53 $\pm$ 0.76\\
\hline
Oracle(9)&0.86 $\pm$ 0.07& 79.11 $\pm$ 1.49 &9.87 $\pm$ 0.52 &3.86 $\pm$ 0.28 &84.79 $\pm$ 0.95\\
\hline
\hline
PV&\cellcolor{green!30}{1.19 $\pm$ 0.09}& \cellcolor{green!30}{70.11 $\pm$ 2.02}&\textbf{9.97 $\pm$ 0.53}&\cellcolor{green!30}{4.22 $\pm$ 0.29}&\cellcolor{green!30}{77.47 $\pm$ 1.18}\\
\hline
MV&\cellcolor{green!30}{1.26 $\pm$ 0.09}& \cellcolor{green!30}{68.88 $\pm$ 2.03}&10.34 $\pm$ 0.52&\cellcolor{green!30}{4.44 $\pm$ 0.29}&\cellcolor{green!30}{77.04 $\pm$ 1.17}\\
\hline
STAPLE&\cellcolor{green!30}{2.90 $\pm$ 0.29}&\cellcolor{green!30}{62.54 $\pm$ 1.99}&\cellcolor{green!30}{21.77 $\pm$ 0.98}&\cellcolor{green!30}8.88 $\pm$ 0.73&\cellcolor{green!30}{73.83 $\pm$ 1.13}\\
\hline
MABMIS&\cellcolor{green!30}1.24 $\pm$ 0.09 &\cellcolor{green!30}68.66 $\pm$ 2.07 &10.07 $\pm$ 0.52&\cellcolor{green!30}4.36 $\pm$ 0.28&\cellcolor{green!30}{77.89 $\pm$ 1.22}\\
\hline
JLF&\cellcolor{green!30}1.15 $\pm$ 0.08&\cellcolor{green!30}72.44 $\pm$ 1.73 & 10.08 $\pm$ 0.54& \cellcolor{green!30}4.30 $\pm$ 0.29&\cellcolor{green!30}{79.50 $\pm$ 1.12}\\
\hline
PB&\cellcolor{green!30}1.31 $\pm$ 0.09& \cellcolor{green!30}67.00 $\pm$ 2.01&10.28 $\pm$ 0.50&\cellcolor{green!30}4.54 $\pm$ 0.30 &\cellcolor{green!30}{77.60 $\pm$ 1.16}\\
\hline
U-Net&\cellcolor{green!30}1.10 $\pm$ 0.12 &\cellcolor{green!30}74.69 $\pm$ 2.15&\cellcolor{green!30}13.79 $\pm$ 3.86& \cellcolor{green!30}4.20 $\pm$ 0.49&80.46 $\pm$ 1.29\\
\hline
VoteNet& 1.02 $\pm$ 0.08& \cellcolor{green!30}75.32 $\pm$ 1.90& 10.89 $\pm$ 0.55&3.84 $\pm$ 0.28& 80.55 $\pm$ 1.13 \\
\hline
VoteNet + U-Net &\textbf{0.99 $\pm$ 0.08} &\textbf{76.23 $\pm$ 1.83} &11.67 $\pm$ 1.48&\textbf{3.78 $\pm$ 0.28}&\textbf{80.75 $\pm$ 1.11}\\
\hline
\end{tabular} 
\end{adjustbox}
\caption{Evaluation metrics for LPBA40 segmentation performance. %Oracle(1, 3, 6, 9) indicate the oracle label fusion results based on the correct label for at least 1(3 /6 /9) of the warped atlas labels.
  \textbf{Avg. Surf. Dist.}: symmetric average surface distance in mm between each segmentation label and the true segmentation. \textbf{Surf. Dice}: Dice score of segmented label surface and true label surface at a tolerance of 1 mm. \textbf{Hausdorff. Dist.}: Hausdorff distance between segmented label volume and true label volume. \textbf{95\% Max. Dist.}: 95 percentile of the maximum symmetric distance between segmented label volume and true label volume. \textbf{Volume Dice}: Average Dice score over segmented labels (excluding the background). We use a Mann-Whitney U-test to check for significant differences to VoteNet+U-Net. We use a significance level of 0.05 and the Benjamini/Hochberg correction~\cite{benjamini1995} for multiple comparisons with a false discovery rate of 0.05. Results are highlighted in green if Vote+U-Net performs significantly better than the corresponding method. VoteNet is always better than U-Net and VoteNet+U-Net achieves the best performance.}
\label{tab:metrics}
\vspace{-0.5ex}
\end{table*}

\noindent
\textbf{Oracle label fusion:} MAS depends on the interplay of image registration and label fusion. Conceivably, given a high-quality registration, one should be able to obtain a high-quality segmentation. To assess how well an ideal label fusion strategy could work, we investigate the behavior of an oracle label fusion method following our Quicksilver atlas to target image registrations. Specifically, Oracle($n$) assigns the correct label to a voxel if at least $n$ warped atlases (out of our 17) correctly label this voxel; otherwise the background label (0) is assigned.\\

\noindent
\textbf{Results:} We use five measures to evaluate segmentations: average surface distance, average surface Dice score (i.e., a surface element is considered overlapping if it is within a certain distance ($\leq$1mm) to the other surface), Hausdorff distance, 95\% maximum surface distance, and average volume Dice score.\\

\noindent\emph{Oracle results:} Tab.~\ref{tab:metrics} shows that Dice scores of Oracle(1) are close to 100\%, indicating that at least one warped atlas label image locally agrees with the manual segmentation. Even Oracle(9) (where the correct label is only assigned if at least 9 of the 17 atlases agree on this labeling) results in Dice scores higher than state-of-the-art approaches. In contrast, MV is significantly worse than Oracle(9). Note that all labels (excluding background) of Oracle(9) are also contained in MV. Hence, MV contains \emph{incorrect} labels for which the majority of atlases \emph{agree}. %\mnl{What is the intuition here; if this is more than half the atlases why is this not the performance that MV achieves; this seems VERY strange.\zpd{(It is high because Oracle only consider the correct label and exclude the wrong label. MV include results that may have high agreement but indeed is wrong. This will greatly influence final result. The intuition here is to tell the readers that predicting which voxel is correct and which is wrong is very important.)}}
Therefore, if VoteNet can identify good subsets of these atlases, a good segmentation should be achievable by majority or plurality voting.\\

\begin{figure}[!t]
  \includegraphics[width=\linewidth]{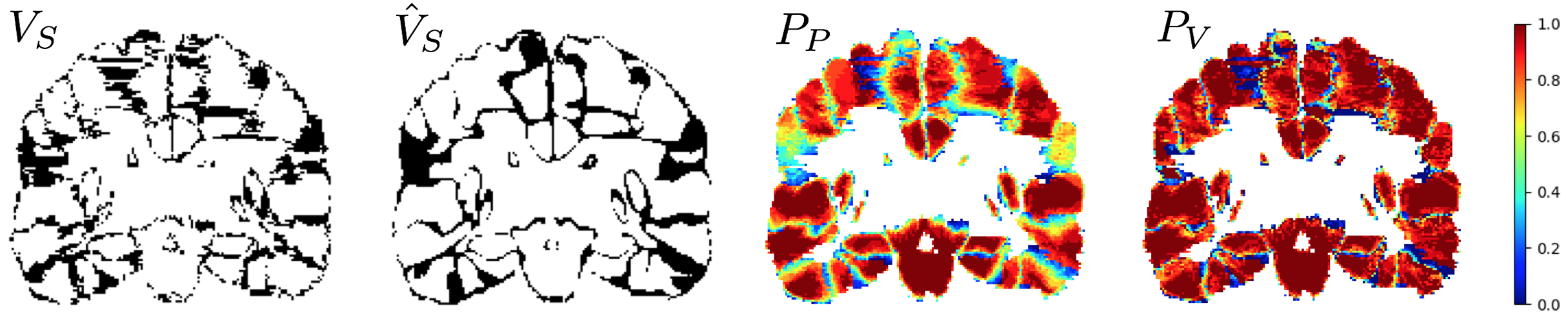}
  \caption{One example of the output of VoteNet and its improvement over plurality voting. \textbf{Left two:} \textbf{$V_S$:} difference between true target image label and warped atlas label \emph{for one atlas}; \textbf{$\hat{V}_S$:} predicted difference from VoteNet. Black indicates that labels are different, white indicates agreement. VoteNet prediction captures most of the label differences. \textbf{Right two:} \textbf{$P_P$:} recall map (i.e., probability of true positives out of all atlases) for plurality voting. \textbf{$P_V$:} recall map of VoteNet. VoteNet greatly improves recall, because it filters out true negatives so that the final voting is more accurate.}
  \label{fig:prob}
\end{figure}

\begin{figure}[t]
  \centering
  \includegraphics[width=0.9\linewidth]{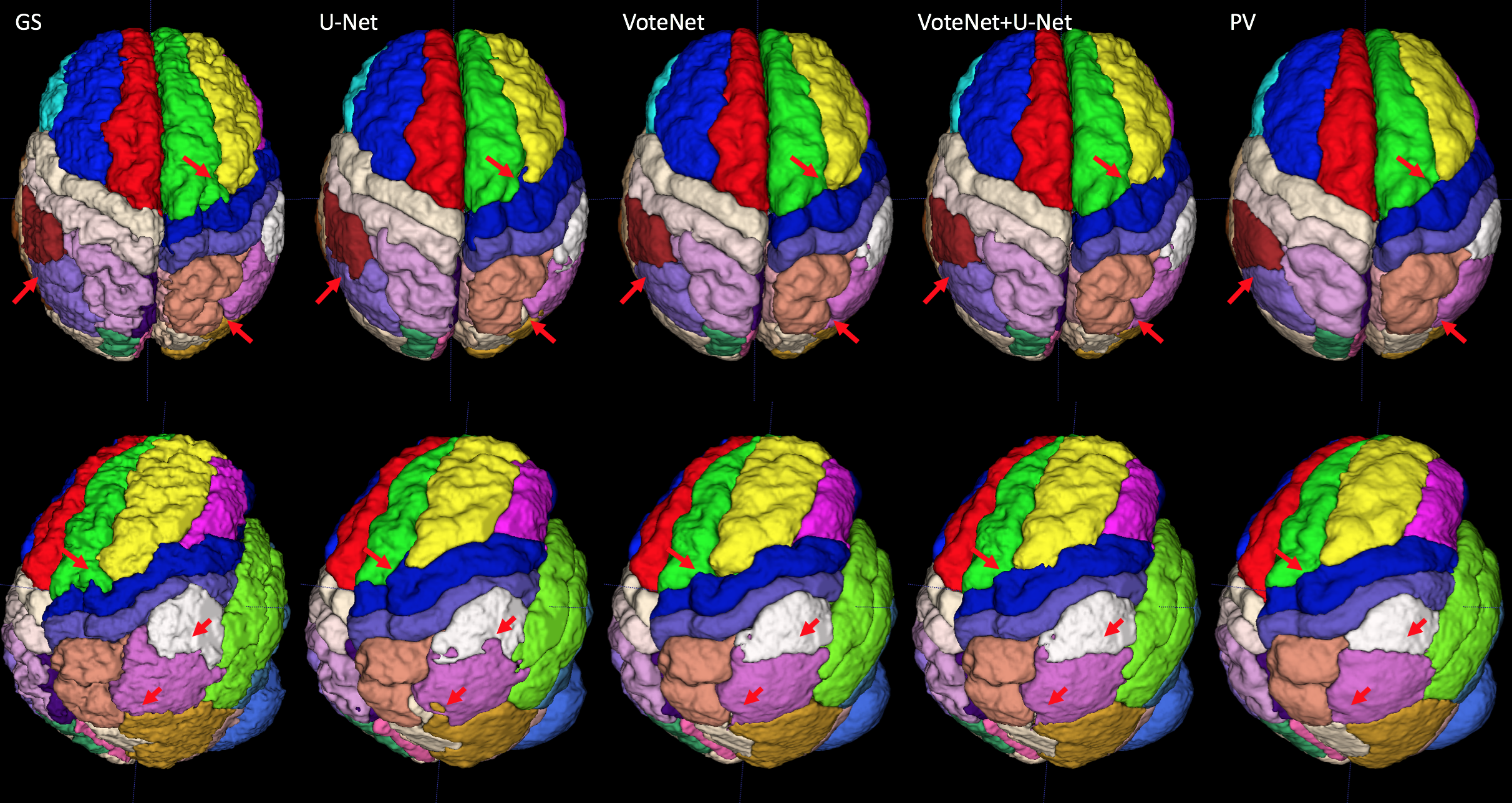}
  \caption{3D rendering of segmented results. Compared with the manual segmentation (GS) U-Net results show shape inconsistencies, highlighted by the red arrows. VoteNet and VoteNet+U-Net retain spatial consistency. Plurality voting (PV) results smooth out cortical structures, but these are well maintained by VoteNet and VoteNet+U-Net.}
  \label{fig:diff}
\end{figure}

\noindent\emph{Label fusion results:} Tab.~\ref{tab:metrics} shows that VoteNet greatly improves performance over MV/PV and significantly outperforms all other evaluated label fusion strategies on most measures. These results also illustrate that VoteNet successfully locally eliminates atlases that would otherwise have tipped the results to incorrect PV label assignments. Further, we observed that there are some voxels ($\approx 7.5\%$ inside the brain) that are not assigned any labels by VoteNet (i.e., locally all warped atlases are rejected). %\mnl{This is a lot.\zpd{(Yes, it is a lot. But I am happy because this means further research can be done here:))}}
We therefore propose a combined VoteNet + U-Net strategy which fills in missing voxels via the U-Net segmentation.  This strategy outperforms both VoteNet and U-Net. %Furthermore, as a U-Net does not have a direct control over spatial segmentation consistency it sometimes result in segmentation inconsistencies. 
There is still a large gap between our VoteNet and the Oracle results. Hence, there is significant room for future improvement. Fig.~\ref{fig:prob} illustrates the performance of VoteNet. The predicted binary mask $\hat{V}_S$ is close to the ground truth binary mask $V_S$, indicating that VoteNet captures most areas of poor label alignments for a given atlas image. In fact, VoteNet achieves a volume Dice score of 0.86 on local atlas selection. Fig.~\ref{fig:prob}(right) also shows that by only retaining locally trustworthy atlases the percentage of true positives (after VoteNet atlas selection)%\mnl{Is this a true positive in the sense of the correct label or for the correct VoteNet prediction. Not totally clear to me.\zpd{(true positive for the correct label after VoteNet prediction filtering. I guess it may confuse people because true positive for $P_M$ is for the correct label while for $P_V$ is for the correct label after VoteNet prediction filetring. After VoteNet, a few voxels are removed. The number of true atlas label left over all left atlases)}}
over all atlases grows significantly. Consequentially, subsequent plurality voting better predicts the correct labels.\\

\noindent\emph{U-Net results:} U-Net results are generally good with respect to the volumetric Dice scores. However, as indicated by the surface measures (in particular, Hausdorff and 95\% maximum surface distance), shapes of segmented structures may locally be distorted, as shape constraints are not straightforward to integrate into a CNN. This drawback is much less present in MAS, as a good deformable image registration method will preserve local structure and topology in target image space (based on transformation smoothness). Fig.~\ref{fig:diff} illustrates this effect. As highlighted by the red arrows, U-Net results often show inconsistent shapes, while VoteNet and VoteNet+U-Net produce shapes more consistent with the manual segmentations. Furthermore, our VoteNet and VoteNet+U-Net retain the cortical foldings, while PV tends to flatten them. This indicates that our proposed approach indeed complements a label fusion method such as PV well. 

%This shape inconsistency is always ignored by researchers seeking to maximize volume dice score with CNNs. In practice, this shape inconsistency can misguide doctors to draw a wrong conclusion on disease progression. Thus, shape information should be a measure when evaluating the segmentation performance. 

%U-Net has a good volume dice score, it can potentially give inconsistent structure segmentation.

%This created a better result, VoteNet+U-Net, than VoteNet alone. Statistical analysis in Tab.~\ref{tab:metrics} show that the combined result outperforms all the other methods (excluding VoteNet) significantly in almost all metrics.

%This illustrates that the idea of filtering out unwanted voxels in warped atlas labels has an effect on correctly some voxel labels that were originally assigned wrongly by MV. Moreover, our proposed method is better than all the other label fusion methods in almost all the metrics. This proves that the idea of excluding wrong voxel labels in MAS is as important as finding a good way to utilize all voxel information. 

%Motivated by this observation, we proposed the VoteNet that can help us predict whether the warped label is correct or not voxel-wisely. The only assumption that we put in this model is that the registration method is robust. More specifically, atlas image to target image give the same alignment if local shape structure is the same. This can be achieved by using Quicksilver~\cite{quicksilver} because it is a patch based method.

\section{Conclusion and Future Work}
\label{sec:conclusion}
We presented a new label fusion method (VoteNet) which helps \emph{locally} select the most trustworthy atlases. With VoteNet, we achieve state-of-the-art segmentation performance, even surpassing a deep network (U-Net) while maintaining spatial shape consistency. We also provided an empirical analysis of best possible achievable performance of our approach, indicating that there is still substantial room for further performance improvements. In particular, it would be interesting to combine VoteNet with more advanced label fusion strategies than plurality voting. As such strategies have shown improved performance for MAS, it is conceivable that they could also further improve our approach, for example, by leveraging local image information to assess atlas to target image similarity. It would also be valuable to explore more advanced network architectures as well as end-to-end formulations integrating the registration network.\\

%. Since the simple majority voting works well, we expect an advanced and better label fusion method embedded into VoteNet could further improve the segmentation performance. Another interesting direction is to transform the whole framework into an end-to-end procedure. 
%For future work, exploring different deep network structures of the VoteNet could be one way to improve the final segmentation performance. Besides, it

\noindent
{\bf Acknowledgements} Research reported in this work was supported by the National Institutes of Health (NIH) and  the  National  Science  Foundation  (NSF)  under  award numbers  NSF  EECS-1711776  and  NIH  1R01AR072013. The content is solely the responsibility of the authors and does not necessarily represent the official views of the NIH or the NSF.

\bibliographystyle{splncs03}
\bibliography{reference/allBibVeryShort}

\end{document}